\documentclass{article}

\usepackage{todonotes}
\usepackage{subcaption}
\usepackage{multirow}

\usepackage[final,nonatbib]{nips_2018}

\usepackage[utf8]{inputenc} 
\usepackage[T1]{fontenc}    
\usepackage{hyperref}       
\usepackage{url}            
\usepackage{booktabs}       
\usepackage{amsfonts}       
\usepackage{nicefrac}       
\usepackage{microtype}      
\usepackage{graphicx}
\usepackage[sort,numbers]{natbib}

\title{Predicting Blood Pressure Response to Fluid Bolus Therapy Using Attention-Based Neural Networks for Clinical Interpretability}

\author{
  Uma M. Girkar\textsuperscript{1}, Ryo Uchimido\textsuperscript{2}, Li-wei H. Lehman\textsuperscript{3} \\
  \textbf{Peter Szolovits\textsuperscript{1}, Leo Celi\textsuperscript{3}, and Wei-Hung Weng\textsuperscript{1}} \\
  \textsuperscript{1}Computer Science and Artificial Intelligence Laboratory, MIT, Cambridge, MA, USA \\
  \textsuperscript{2}Emergency Department, Beth Israel Deaconess Medical Center, Boston, MA, USA \\
  \textsuperscript{3}Laboratory for Computational Physiology, MIT, Cambridge, MA, USA \\
  \texttt{ruchimido@bidmc.harvard.edu} \\
  \texttt{\{umag, lilehman, psz, lceli, ckbjimmy\}@mit.edu}\\
}

\begin{document}

\maketitle

\begin{abstract}
Determining whether hypotensive patients in intensive care units (ICUs) should receive fluid bolus therapy (FBT) has been an extremely challenging task for intensive care physicians as the corresponding increase in blood pressure has been hard to predict.
Our study utilized regression models and attention-based recurrent neural network (RNN) algorithms and a multi-clinical information system large-scale database to build models that can predict the successful response to FBT among hypotensive patients in ICUs.
We investigated both time-aggregated modeling using logistic regression algorithms with regularization and time-series modeling using the long short term memory network (LSTM) and the gated recurrent units network (GRU) with the attention mechanism for clinical interpretability.
Among all modeling strategies, the stacked LSTM with the attention mechanism yielded the most predictable model with the highest accuracy of 0.852 and area under the curve (AUC) value of 0.925. 
The study results may help identify hypotensive patients in ICUs who will have sufficient blood pressure recovery after FBT.

\end{abstract}

\section{Introduction}
Excessive positive fluid balance in critically ill patients in the intensive care units (ICUs) has been proposed as a major risk factor for severe organ dysfunctions, prolonged mechanical ventilation, longer length of stay in ICU, and increased mortality~\citep{payen2008positive,lee2015association,vincent2006sepsis}. 
Fluid bolus therapy (FBT), the rapid infusion of fluid, has been recommended as the primary-line treatment for acute hypotensive episode (AHE) that occurs in about 41\% of patients in ICU~\citep{saeed2002mimic}. However, previous studies have reported that approximately one-third of the acute hypotensive patients do not successfully respond to FBT treatment~\citep{natalini2012prediction,lammi2015response,garcia2015effects}.Considering that FBT accounts for about 30-50\% of the total fluid volume administrated for ICU patients~\citep{bihari2013post}, avoiding the administration of FBT that will not successfully resolve AHE might prevent an inappropriate increase of the total fluid volume administrated to ICU patients~\citep{lammi2015response}.

The prediction of whether ICU patients respond to FBT has been studied for decades. Previous studies have focused on predicting cardiac output as patients' response to FBT since fluid infusion increases cardiac output by increasing the venous return to heart~\citep{toscani2017impact}. A 15\% increase in cardiac output after FBT administration has been used as the success benchmark of FBT~\citep{glassford2014physiological}. However, a recent clinical observational study, the FENICE study, reported that while only 11\% of ICU clinicians use the increased cardiac output as the patient's positive response to FBT, 67\% of them judged the patient's response as positive when the patient's blood pressure increases~\citep{cecconi2015fluid}. Although cardiac output is a gold-standard measure to evaluate patient's physiological response to FBT based on the result of the FENICE study, the blood pressure response can be used as a practical alternative to evaluate the patient's response to FBT in the ICU.

In this study, we first applied logistic regression models and attention-based recurrent neural network algorithms with time-structured data to develop models for predicting successful blood pressure response to the first FBT administered to critically ill patients during their first 24 hours in the ICU.
A previous study had used non-time structured data without machine learning algorithms to construct the models to predict blood pressure response to FBT, yet the performance of the area under the curve of the receiver operating characteristic (AUC) ranged from 0.5 to 0.6, which was not acceptable to be implemented in the clinical setting~\citep{natalini2012prediction}. 
Therefore, we investigated both time-aggregated and time-series structured data for modeling.
Regularized logistic regression, as well as the stacked long short term memory network (LSTM) and gated recurrent units network (GRU) models with and without the attention mechanism were applied to identify hypotensive critically ill patients in the ICU who will obtain sufficient blood pressure recovery after the FBT~\citep{hochreiter1997long,cho2014properties,bahdanau2014neural}. 

The goal of this study is to achieve high model performance and clinical interpretability for real-world implementation.
Particularly, the contributions of this study include the following:
\begin{enumerate}
    \item This is the first study that utilizes machine learning algorithms to develop models for predicting successful blood pressure response to FBT in critically ill patients in the ICU
    \item The regularized regression model and LSTM/GRU models with the attention mechanism provide us with certain important features for clinical interpretability. 
\end{enumerate}

\section{Methods}
\paragraph{Dataset and Cohort}
Study data was collected from the MIMIC-III database~\citep{johnson2016mimic}, which contains 58,976 ICU patients admitted to the Beth Israel Deaconess Medical Center (BIDMC), a large, tertiary medical center in Boston, Massachusetts, USA. 
The database contains detailed information on patients admitted between 2001 and 2012, including hospital administrative data, vital signs, medications, laboratory test results and survival data after hospital discharge.
For the cohort selection, we considered only (1) the first ICU stay during the hospital stay, (2) patients who were more than 18 years old on the first day of admission, (3) patients with a length of ICU stay more than 12 hours in order to include only true ICU patients, (4) patients who received their first FBT during their first 24 hours in the ICU, where FBT is defined as the crystalloid fluid infusion rate $>$248 ml/hr and volume $>$248 ml
, and (5) patients who are hypotensive (mean atrial pressure (MAP) =< 65 mmHg)  when the first FBT started. 
17,977 patients were selected for the final patient cohort.

\paragraph{Clinical Covariates/Features}
We extracted 29 clinically meaningful features from the MIMIC-III database, which include time-static features: (1) patient demographics (age, gender, race/ethnicity, weight, height, and SOFA score at ICU admission) and (2) comorbidity condition using Elixhauser coding algorithm, 
and time-varying features: (1) physiological parameters (heart rate, respiratory rate, temperature, oxygen saturation, systolic blood pressure, diastolic blood pressure, mean arterial pressure,  and urine output), (2) laboratory examination results (pH, PaO2, PaCO2, bicarbonate, base excess, lactate, sodium, potassium and chloride), and (3) vasopressor dosage measured (norepinephrine, epinephrine, phenylephrine, vasopressin and dopamine).
For time-aggregated modeling, we collected the laboratory examination results, physiological parameters and vasopressor dosage information, except MAP, at the time interval between 30 minutes before and 30 minutes after 3 important time points --- six hours before the first FBT, two hours before the first FBT, and right at the start of the first FBT.
For time-series modeling, all values of time-varying features before FBT were extracted along with their time information.

All raw values of features were normalized in population level and re-scaled to obtain a value between zero and one.
The missing values were imputed by median values of the features. 

\paragraph{Clinical Outcome}
The primary outcome of this study is the physiological response, which is reflected by the change of MAP. 
The outcome variable is binary, either success or failure.
The successful FBT is defined by intensive care experts as the presence of $max(\mathrm{MAP}_{fbt}) > 1.15 \times average(\mathrm{MAP}_{all})$ at least once, where $max(\mathrm{MAP}_{fbt})$ is the maximal MAP from the FBT starting time to two hours after FBT, and $average(\mathrm{MAP}_{all})$ is the average MAP from 30 minutes before FBT until 10 minutes after FBT. 

\subsection{Experiment Settings}
We divided all patients in our cohort into a 75\% training and 25\% testing set, and further divided the training set into a 75\%/25\% split for cross-validated model evaluation. 

Two categories of experiment settings were performed---time-aggregated setting and time-series setting. 
For both settings, the normalized raw feature representation and the distributed representation were investigated. 
We used a stacked autoencoder to generate the distributed representation, which was joint-trained with the classifier and optimized using the reconstruction loss by the Adam optimizer.
To reduce the dimensionality of raw features, we used 32-dimension representation for the time-aggregated setting. and 25-dimension representation for the time-series setting.

\paragraph{Time-Aggregated Setting}
We used logistic regression with L1-regularization (LASSO regression), logistic regression with L2-regularization (ridge regression), and the multiple layer perceptron (MLP) model. 
The $\mathsf{liblinear}$ solver supported L1-regularization and the $\mathsf{lbfgs}$ solver supported L2-regularization~\citep{lin2008liblinear,okazaki2010liblbfgs}. 
For the MLP model, we used two hidden layers both with 64 neurons, ReLU activation and 50\% of dropout and optimized a binary cross entropy loss by the RMSprop optimizer.
We output a probabilistic prediction of the target for all models.

\paragraph{Time-Series Setting}
All features were transformed into a time-series data with different time resolutions by collecting data at equal interval lengths every certain number of minutes based on the number of timesteps specified. For non time-varying covariate variables, the same value was recorded for all timesteps. For time-varying covariate variables, data collection started six hours before the first FBT was administered and ended at the time when the first FBT was administered.   
We used 12, 36, and 72 timesteps for sequential modeling. 
We adopted the stacked LSTM and the corresponding GRU models, and also the above models with attention mechanism for prediction.
We used three layers for the LSTM/GRU models with 32 dimensions, and output the probabilistic prediction of the softmax layer. 
The models were optimized through a binary cross-entropy loss by the RMSprop optimizer. 

\paragraph{Evaluation}
For the binary classification task, we computed the accuracy and AUC of all models. 
We also identified the important features in the LASSO regression (time-aggregated setting) and in the LSTM/GRU with attention mechanism (time-series setting) for human experts to qualitatively evaluate the interpretability of the models.
The top five features with the highest coefficients (absolute value) in LASSO regression, and the most important timesteps with the highest weights in the neural network models with attention mechanism were extracted for interpretation.

\section{Results}
The results of the binary classification task of predicting whether the FBT yielded successful blood pressure improvement using the different machine learning settings are shown in Table~\ref{table:res}. 
In general, the attention-based neural network models which considered time sequence with higher temporal granularity yielded higher performance for prediction. 
Using the distributed representations yield comparable results to using the raw features. 
The result indicates that the autoencoder-derived representation is compact but still informative even if it is in a lower dimension.
This approach may be valuable when we scale up the method with more covariates, or even when applying massive numbers of features from other modalities.

\begin{table}[h!]
    \centering\footnotesize
    \setlength{\tabcolsep}{4pt}
    \begin{tabular}{*{6}{c}}
        \toprule
        \multirow{2}[2]{*}{Algorithm}
        & \multirow{2}[2]{*}{Timesteps}
        & \multicolumn{2}{c}{Accuracy}
        & \multicolumn{2}{c}{AUC}
        \\ \cmidrule(lr){3-4} \cmidrule(lr){5-6}
        &
        & Raw features
        & Distributed
        & Raw features
        & Distributed 
        \\
        \midrule
        L1-regularized logistic regression & - & 0.706 & 0.688 & 0.680 & 0.659 \\
        L2-regularized logistic regression & - & 0.699 & 0.695 & 0.679 & 0.664 \\
        Multiple layer perceptron & - & \textbf{0.712} & 0.688 & \textbf{0.690} & 0.642 \\
        \midrule
        LSTM & 12 & 0.751 & 0.705 & 0.818 & 0.718 \\
        GRU & 12 & 0.748 & 0.721 & 0.813 & 0.770 \\
        LSTM + Attention & 12 & 0.747 & 0.727 & 0.822 & 0.795 \\
        GRU + Attention & 12 & 0.747 & 0.716 & 0.818 & 0.786 \\
        LSTM & 36 & 0.814 & 0.827 & 0.899 & 0.899 \\
        GRU & 36 & 0.820 & 0.794 & 0.898 & 0.869 \\
        LSTM + Attention & 36 & 0.819 & 0.820 & 0.902 & 0.895 \\
        GRU + Attention & 36 & 0.812 & 0.777 & 0.893 & 0.858 \\
        LSTM & 72 & 0.843 & 0.834 & \textbf{0.926} & 0.915 \\
        GRU & 72 & 0.848 & 0.836 & 0.922 & 0.904 \\
        LSTM + Attention & 72 & \textbf{0.852} & 0.831 & 0.925 & 0.920 \\
        GRU + Attention & 72 & 0.841 & 0.803 & 0.917 & 0.882 \\
        \bottomrule
    \end{tabular}
    \vspace{5pt}
    \caption{Model performance in accuracy and AUC between different experimental settings. \textbf{Boldface} denotes the best performance in each group.}
    \label{table:res}
    \vspace{-10pt}
\end{table}

The top five important features learned from LASSO regression include the patient's respiratory rate, diastolic pressure, temperature, and bicarbonate and base excess levels in blood. 
These features have been little considered as having predictive value in previous clinical studies since there is still no consensus for this critical problem.
Further investigation from clinical sites should be done to prove these features are clinically and physiologically plausible.  

\begin{figure}
    \begin{subfigure}{0.9\textwidth}
        \includegraphics[width=1.05\linewidth]{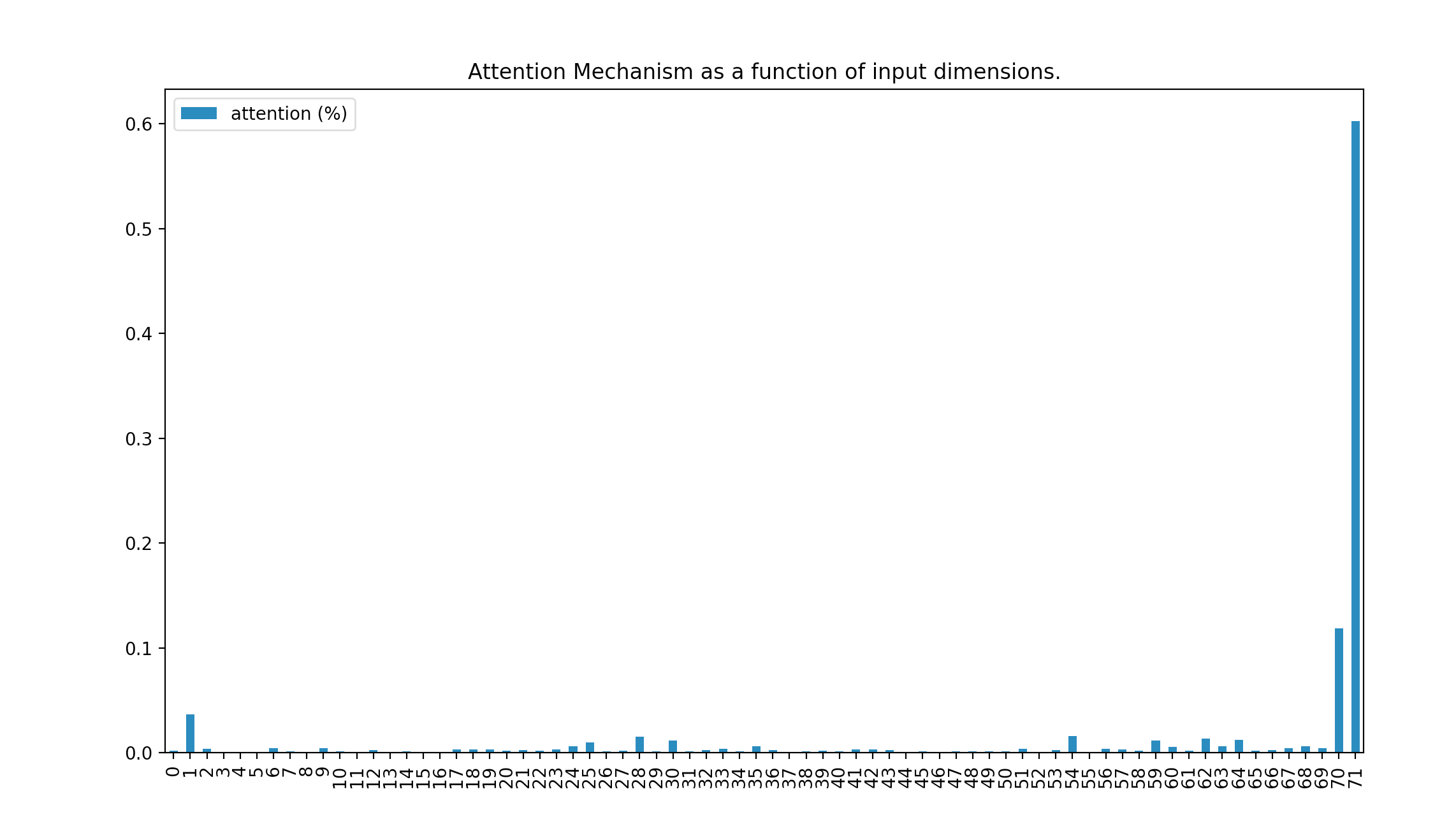}
        \label{subfig:attn}
    \end{subfigure}
    \caption{The interpretability of the neural network model through attention mechanism. The result is the average attention weight of 300 input patient cases in each timestep until FBT. The timesteps closer to FBT have higher impact to the prediction.}
    \label{fig:joint}
    \vspace{-10pt}
\end{figure}

We are also able to identify the key timesteps using the recurrent neural network models (LSTM/GRU) with the attention mechanism by extracting the attention weights. 
Figure~\ref{fig:joint}. demonstrates that the attention mechanism captured the key time points for model decision through computing the attention weights of each timestep. 
The result is clinically meaningful since the time points closer to the time of FBT are the most important, which is explainable from the clinical perspective.

\section{Conclusions}
In summary, we first demonstrate the capability of the LSTM and GRU networks to predict the blood pressure response to FBT in hypotensive, critically ill patients. 
We found that the stacked LSTM and GRU recurrent neural networks with attention mechanism yielded the most predictable models based on the metrics of accuracy and AUC. 
Clinical interpretability is also achieved by identifying the important timesteps in the time-series setting. In the time-aggregated setting, we identified novel features that may provide new clues to the prediction of blood pressure response to FBT. 

Some limitations in the study provide possibilities for future directions. 
We will investigate the optimal strategy to determine when and how much fluid bolus a patient should receive using reinforcement learning.
Furthermore, we will improve on the patient state representation to potentially perform transfer learning tasks, or add other features to better perform data modeling. 
On the clinical side, we will include (1) the dosage values of analgesic and sedative drugs administered to patients, and (2) the parameters of mechanical ventilation settings as covariates to make the model more robust. 
Finally, we hope to develop a generalizable model that can be used to predict the success of FBT on any given day that the patient is admitted to the ICU.
The study results may support intensive care clinicians to identify whether the hypotensive episode in ICU patients will resolve with FBT.



\bibliographystyle{plain}
\bibliography{main}

\begin{thebibliography}{10}

\bibitem{bahdanau2014neural}
Dzmitry Bahdanau, Kyunghyun Cho, and Yoshua Bengio.
\newblock Neural machine translation by jointly learning to align and
  translate.
\newblock {\em ICLR}, 2015.

\bibitem{bihari2013post}
Shailesh Bihari, Shivesh Prakash, and Andrew~D Bersten.
\newblock Post resusicitation fluid boluses in severe sepsis or septic shock:
  prevalence and efficacy (price study).
\newblock {\em Shock}, 40(1):28--34, 2013.

\bibitem{cecconi2015fluid}
Maurizio Cecconi, Christoph Hofer, Jean-Louis Teboul, Ville Pettila, Erika
  Wilkman, Zsolt Molnar, Giorgio Della~Rocca, Cesar Aldecoa, Antonio Artigas,
  Sameer Jog, et~al.
\newblock Fluid challenges in intensive care: the fenice study.
\newblock {\em Intensive care medicine}, 41(9):1529--1537, 2015.

\bibitem{cho2014properties}
Kyunghyun Cho, Bart Van~Merri{\"e}nboer, Dzmitry Bahdanau, and Yoshua Bengio.
\newblock On the properties of neural machine translation: Encoder-decoder
  approaches.
\newblock {\em Eighth Workshop on Syntax, Semantics and Structure in
  Statistical Translation (SSST-8)}, 2014.

\bibitem{lin2008liblinear}
Rong-En Fan, Kai-Wei Chang, Cho-Jui Hsieh, Xiang-Rui Wang, and Chih-Jen Lin.
\newblock {LIBLINEAR}: A library for large linear classification.
\newblock {\em Journal of Machine Learning Research}, 9:1871--1874, 2008.

\bibitem{garcia2015effects}
Manuel Ignacio~Monge Garc{\'\i}a, Pedro~Guijo Gonz{\'a}lez, Manuel~Gracia
  Romero, Anselmo~Gil Cano, Chris Oscier, Andrew Rhodes, Robert~Michael
  Grounds, and Maurizio Cecconi.
\newblock Effects of fluid administration on arterial load in septic shock
  patients.
\newblock {\em Intensive care medicine}, 41(7):1247--1255, 2015.

\bibitem{glassford2014physiological}
Neil~J Glassford, Glenn~M Eastwood, and Rinaldo Bellomo.
\newblock Physiological changes after fluid bolus therapy in sepsis: a
  systematic review of contemporary data.
\newblock {\em Critical care}, 18(6):696, 2014.

\bibitem{hochreiter1997long}
Sepp Hochreiter and J{\"u}rgen Schmidhuber.
\newblock Long short-term memory.
\newblock {\em Neural computation}, 9(8):1735--1780, 1997.

\bibitem{johnson2016mimic}
Alistair~EW Johnson, Tom~J Pollard, Lu~Shen, H~Lehman Li-wei, Mengling Feng,
  Mohammad Ghassemi, Benjamin Moody, Peter Szolovits, Leo~Anthony Celi, and
  Roger~G Mark.
\newblock {MIMIC-III}, a freely accessible critical care database.
\newblock {\em Scientific data}, 3:160035, 2016.

\bibitem{lammi2015response}
Matthew~R Lammi, Brianne Aiello, Gregory~T Burg, Tayyab Rehman, Ivor~S Douglas,
  Arthur~P Wheeler, National~Institutes of~Health, ARDS~Network Investigators,
  et~al.
\newblock Response to fluid boluses in the fluid and catheter treatment trial.
\newblock {\em Chest}, 148(4):919--926, 2015.

\bibitem{lee2015association}
Joon Lee, Emma de~Louw, Matthew Niemi, Rachel Nelson, Roger~G Mark, Leo~Anthony
  Celi, Kenneth~J Mukamal, and John Danziger.
\newblock Association between fluid balance and survival in critically ill
  patients.
\newblock {\em Journal of internal medicine}, 277(4):468--477, 2015.

\bibitem{natalini2012prediction}
Giuseppe Natalini, Antonio Rosano, Carmine~Rocco Militano, Antonella Di~Maio,
  Pierluigi Ferretti, Michele Bertelli, Federica de~Giuli, and Achille
  Bernardini.
\newblock Prediction of arterial pressure increase after fluid challenge.
\newblock {\em BMC anesthesiology}, 12(1):3, 2012.

\bibitem{okazaki2010liblbfgs}
Naoaki Okazaki and J~Nocedal.
\newblock Liblbfgs: a library of limited-memory
  broyden-fletcher-goldfarb-shanno (l-bfgs).
\newblock {\em URL http://www.chokkan.org/software/liblbfgs}, 2010.

\bibitem{payen2008positive}
Didier Payen, Anne~Corn{\'e}lie de~Pont, Yasser Sakr, Claudia Spies, Konrad
  Reinhart, and Jean~Louis Vincent.
\newblock A positive fluid balance is associated with a worse outcome in
  patients with acute renal failure.
\newblock {\em Critical care}, 12(3):R74, 2008.

\bibitem{saeed2002mimic}
Mohammed Saeed, Christine Lieu, Greg Raber, and Roger~G Mark.
\newblock {MIMIC II}: a massive temporal icu patient database to support
  research in intelligent patient monitoring.
\newblock In {\em Computers in Cardiology, 2002}, pages 641--644. IEEE, 2002.

\bibitem{toscani2017impact}
Laura Toscani, Hollmann~D Aya, Dimitra Antonakaki, Davide Bastoni, Ximena
  Watson, Nish Arulkumaran, Andrew Rhodes, and Maurizio Cecconi.
\newblock What is the impact of the fluid challenge technique on diagnosis of
  fluid responsiveness? a systematic review and meta-analysis.
\newblock {\em Critical Care}, 21(1):207, 2017.

\bibitem{vincent2006sepsis}
Jean-Louis Vincent, Yasser Sakr, Charles~L Sprung, V~Marco Ranieri, Konrad
  Reinhart, Herwig Gerlach, Rui Moreno, Jean Carlet, Jean-Roger Le~Gall, and
  Didier Payen.
\newblock Sepsis in european intensive care units: results of the {SOAP} study.
\newblock {\em Critical care medicine}, 34(2):344--353, 2006.

\end{thebibliography}

\end{document}